\documentclass[10pt,twocolumn,letterpaper]{article}

\usepackage[final]{cvpr}      
\usepackage{tabularx}
\usepackage{times}
\usepackage{caption}
\usepackage{subcaption}
\usepackage{epsfig}
\usepackage{graphicx}
\usepackage{amsmath}
\usepackage{amssymb}
\usepackage{array}
\usepackage{colortbl}
\usepackage{booktabs}
\usepackage{bbm}
\usepackage{multirow}
\usepackage{multicol}
\usepackage{makecell}
\usepackage{xcolor}
\usepackage{xspace}
\usepackage{float}
\usepackage{color}
\usepackage{enumitem}
\usepackage{pifont}

\newcommand{\upscore}[1]{\textcolor{codegreen}{(+\textbf{#1})}}
\newcommand{\downscore}[1]{\textcolor{gray}{(- \textbf{#1})}}

\usepackage{epsfig}
\usepackage{graphicx}
\usepackage{amsmath}
\usepackage{amssymb}

\usepackage{booktabs}
\usepackage{tabulary}
\usepackage{dblfloatfix}
\usepackage{transparent}

\usepackage{algorithm}
\usepackage{listings}
\usepackage{algorithmic}
\usepackage{balance}

\usepackage[margin=4pt,font=small,labelfont=bf,labelsep=endash,tableposition=top]{caption}

\definecolor{codegreen}{rgb}{0.0,0.6,0.0}

\definecolor{codegreen}{rgb}{0,0.5,0}
\definecolor{codeblue}{rgb}{0.25,0.5,0.5}
\definecolor{codegray}{rgb}{0.6,0.6,0.6}

\usepackage[pagebackref=true,breaklinks=true,colorlinks,bookmarks=false]{hyperref}

\usepackage[capitalize]{cleveref}
\crefname{section}{Sec.}{Secs.}
\Crefname{section}{Section}{Sections}
\Crefname{table}{Table}{Tables}
\crefname{table}{Tab.}{Tabs.}

\def\cvprPaperID{0000} 
\def\confName{CVPR}
\def\confYear{9999}
\usepackage{bm}
\newcommand{\myparagraph}[1]{{\vspace{.5em} \noindent \bf #1}}
\newcommand{\app}{\raise.17ex\hbox{$\scriptstyle\sim$}}

\begin{document}

\title{DanceTrack: Multi-Object Tracking in Uniform Appearance and Diverse Motion}

\author
{
Peize Sun$^{1*}$, 
~
Jinkun Cao$^{2*}$, 
~
Yi Jiang$^{3}$,
~
Zehuan Yuan$^{3}$, 
~
Song Bai$^{3}$, 
~
Kris Kitani$^{2}$,
~
Ping Luo$^{1}$
\\[0.2cm]
${^1}$The University of Hong Kong ~~~
${^2}$Carnegie Mellon University~~~
${^3}$ByteDance Inc.
}

\twocolumn[{
\maketitle
\begin{figure}[H]
\hsize=\textwidth
\centering
\vspace{-8mm}
\includegraphics[width=1.0\textwidth]{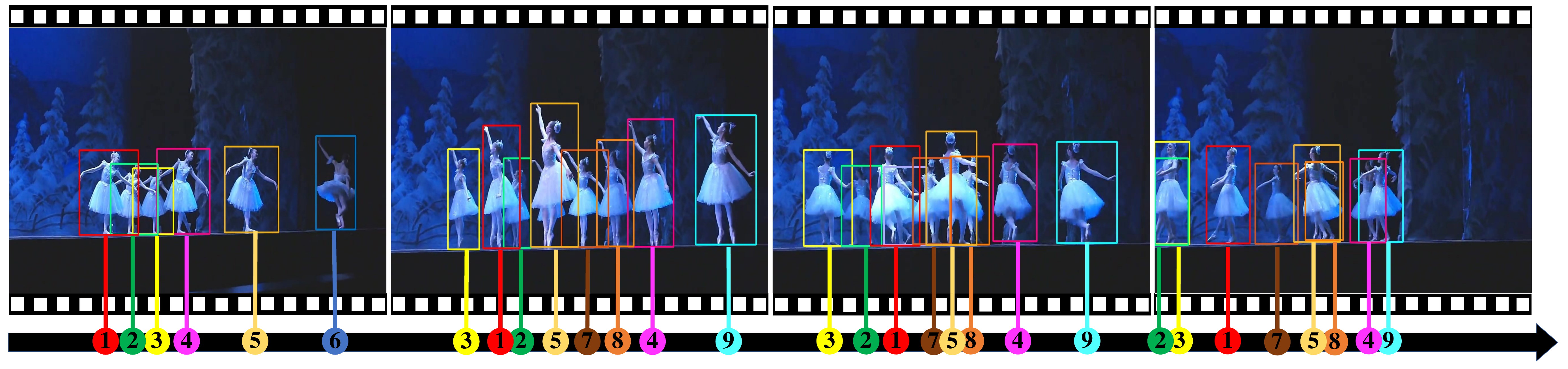}
\caption{Sample images from a video in DanceTrack: 1st, 66th, 307th and 327th frame in DanceTrack0027 video. The emphasized properties of this dataset are (1) \textit{uniform appearance}: humans are in highly similar and almost undistinguished appearance. (2) \textit{diverse motion}: they are in complicated motion and interaction pattern. The numbers below show their identifications which experience frequent relative position switches and occlusions. We expect the combination of uniform appearance and complicated motion pattern makes DanceTrack a platform to encourage more comprehensive and intelligent multi-object tracking algorithms.}
\label{fig:boom}
\vspace{3mm}
\end{figure}
}]

\footnote{* indicates equal contribution.}
\pagestyle{plain}
\thispagestyle{plain}

\vspace{-4mm}
\begin{abstract}
    A typical pipeline for multi-object tracking (MOT) is to use a detector for object localization, and following re-identification (re-ID) for object association. This pipeline is partially motivated by recent progress in both object detection and re-ID, and partially motivated by biases in existing tracking datasets, where most objects tend to have distinguishing appearance and re-ID models are sufficient for establishing associations. In response to such bias, we would like to re-emphasize that methods for multi-object tracking should also work when object appearance is not sufficiently discriminative. To this end, we propose a large-scale dataset for multi-human tracking, where humans have similar appearance, diverse motion and extreme articulation. As the dataset contains mostly group dancing videos, we name it ``DanceTrack''.  We expect DanceTrack to provide a better platform to develop more MOT algorithms that rely less on visual discrimination and depend more on motion analysis. We benchmark several state-of-the-art trackers on our dataset and observe a significant performance drop on DanceTrack when compared against existing benchmarks.
    The dataset, project code and competition is released at:
    {\footnotesize \url{https://github.com/DanceTrack}}.
\end{abstract}

\vspace{-0.5cm}
\section{Introduction}
\vspace{-0.2cm}
Object tracking has been long studied and can be beneficial to applications such as autonomous driving, video analysis and robot planning~\cite{bergmann2019tracking,cao2021instance,yilmaz2006object,rangesh2019no}. Multi-object tracking aims to localize and associate objects of interest along time. Interestingly, we observe that recent developments in multi-object tracking rely heavily on a paradigm of detection followed by re-ID, where mostly appearance cues are used to associate objects. This trend in algorithmic development makes existing solutions fail catastrophically in situations where objects share very similar appearance, \textit{e.g.}, group dancing where performers wear uniform clothes. It inspires us to propose more comprehensive solutions by taking other cues into modeling, such as object motion patterns and temporal dynamics.

As with many other areas of computer vision, the development of multi-object tracking is influenced by benchmark datasets. Based on specified datasets~\cite{MOT16,MOT20,KITTI,BDD}, data-driven methods are sometimes argued to be biased to certain data distributions. 
In this work, we recognize the limitations of existing multi-object tracking datasets lie on that many objects have distinct appearance and the motion pattern of objects are very regular or even linear. Motivated by these dataset properties, most recently developed multi-object tracking methods~\cite{DeepSORT,FairMOT,quasidense,TraDeS} highly rely on appearance matching to associate detected objects while taking little other cues into consideration. The dominant paradigm will fail in situations out of the biased distribution. This phenomenon is not what we expect if we aim to build more general and intelligent tracking algorithms.

To provide a new platform for more comprehensive multi-object tracking studies, we propose a new dataset in this paper. 
Because it mostly contains group dancing videos, we name it ``DanceTrack''. The dataset contains over 100K image frames (almost $10 \times$ more than MOT17 datatset~\cite{MOT16}). As shown in Figure~\ref{fig:boom}, the emphasized properties of this dataset are (1) \textbf{uniform appearance}: people in videos wear very similar or even the same clothes, making their visual features hard to be distinguished by re-ID model and (2) \textbf{diverse motion}: people usually have very large-range motion and complex body gesture variation, proposing higher requirements for motion modeling. The second property also brings occlusion and crossover as a side-effect that human body has a large ratio of overlap with each other and their relative position exchanges frequently.

With the proposed dataset, we build a new benchmark including existing popular multi-object tracking methods. The results prove that current state-of-the-art algorithms~\cite{CenterTrack,FairMOT,quasidense,Transtrack,TraDeS,MOTR,bytetrack} fail to make satisfactory performance when they simply use appearance matching or linear motion models to associate objects across frames. Considering the cases focused on in this dataset happen frequently in our real life, we believe it shows the limitations of existing multi-object tracking algorithms on practical applications. 
To provide potential guidelines for further research, we analyze a range of choices in associating objects and achieve some beneficial conclusions: (1) fine-grained representations of objects, e.g., segmentation and pose, exhibit better ability than coarse bounding box; (2) depth information shows positive influence on associating objects, though we are solving a 2D tracking task; (3) motion modeling of temporal dynamics is important.

To conclude, the key contributions of our work to the object tracking community are as follows:
\begin{enumerate}[leftmargin=*]
    \setlength\itemsep{0.1em}
    \item We build a new large-scale multi-object tracking dataset, DanceTrack, covering the scenarios where tracking suffers from low distinguishability of object appearance and diverse non-linear motion patterns.
    \item We benchmark baseline methods on this newly built dataset with various evaluation metrics, showing the limitation of existing multi-object tracking algorithms. 
    \item We provide comprehensive analysis to discover more cues for developing multi-object trackers that are more robust in complicated real-life situations.
\end{enumerate}
\section{Related Works}
\vspace{-0.2cm}
\myparagraph{Multi-object tracking datasets.}
Many multi-object tracking datasets have been proposed for different scenarios. Similar to our proposed dataset, many existing datasets focus on human tracking. PETS2009~\cite{PETS2009} dataset is one of the earliest in this area. The more recent MOT15~\cite{MOT15}, MOT17~\cite{MOT16} and MOT20~\cite{MOT20} datasets are all popular in this community. These datasets are limited in the aspects of undistinguished appearance and diverse motion. For example, MOT17 contains only a handful of videos and scenarios. Even MOT20 increases the density of objects and emphasizes the occlusion among them, the movements of objects are very regular and they still have distinguishable appearances. Association by pure appearance matching~\cite{quasidense} could easily make success on these datasets and we will show that given the perfect detector, the tracking problem on these datasets can be solved by a very naive association strategy, in Section~\ref{sec:oracle}.

Besides, many other datasets are proposed for diverse objectives, e.g., WILDTRACK~\cite{wildtrack} for multi-camera tracking, Youtube-VIS~\cite{youtubevos} for video instance segmentation and tracking. With the increasing attraction of autonomous driving, some datasets are specifically built where the objects of interest are vehicles and pedestrians. KITTI~\cite{KITTI} is one of the earliest large-scale multi-object tracking datasets for driving scenarios. More recently, BDD100K~\cite{BDD}, Waymo~\cite{waymo} and KITTI360~\cite{KITTI360} are made available to the public, still focusing on autonomous driving scenarios but providing much larger scale data than KITTI. With the limitation of lanes and traffic rules, the motion patterns of objects in these datasets are even more regular than those focusing on only moving people.  There are many datasets focusing on more diverse object categories than persons and vehicles. The ImageNet-Vid~\cite{ImageNet} provides trajectory annotations for 30 object categories in over 1000 videos and TAO~\cite{TAO} annotates even 833 object categories to study object tracking on long-tailed distribution.

\myparagraph{Tracking by matching appearance.}
In the recent development of multi-object tracking, appearance similarity serves as the dominant cue in many popular methods. 
For example, 
JDE~\cite{JDE} and FairMOT~\cite{FairMOT} learn object localization and appearance embedding using a shared backbone for better appearance representation. QDTrack~\cite{quasidense} designs a contrastive training paradigm and dense localization for object detection and uses highly sensitive appearance comparison to match objects across frames. 
More recently, with the new focus of applying transformers~\cite{vaswani2017attention} in vision tasks, TransTrack~\cite{Transtrack}, TrackFormer~\cite{Trackformer} and MOTR~\cite{MOTR} make attempts to leverage the attention mechanism in tracking objects in videos. In these works, the features of previous tracklets are passed to the following frames as the query to associate the same objects across frames. The appearance information contained in the query is critical to keep tracklet consistency. 

Although the rise of deep-learning model brings much more powerful visual representations than ever before, we still witness the failure of appearance matching in many real-world situations and expect to improve the tracking performance by taking other cues into account.

\myparagraph{Motion analysis in object tracking.} 
The displacement of objects-of-interest provides important cues for object tracking. Tracking objects by estimating their motions has inspired a line of researches. These tracking algorithms mainly follow the tracking-by-detection paradigm. 
Sequential analysis tools such as Particle filter~\cite{particle1, particle2} and Kalman filter~\cite{kalman1960new} are found efficient in such applications, for example, SORT~\cite{SORT} is developed on the Kalman filter motion model. 
Even though motion analysis has been used in many object tracking methods~\cite{JDE,FairMOT,bytetrack}, all these methods can only handle simple linear motion pattern and provide limited help in more complicated situations. Furthermore, as deep networks bring the revolutionary ability to extract high-quality visual features, DeepSORT~\cite{DeepSORT} tries to combine deep visual features and motion models to gain performance gain. Since then, motion-based object tracker has shown weak competitiveness and many focuses are towards appearance cues. 

However, we argue that a more comprehensive and intelligent tracking algorithm should pay more attention to motion analysis since appearance is not always reliable. 

\section{DanceTrack}
\subsection{Dataset Construction}
\label{sec:dataset_design}
\vspace{-0.2cm}
\myparagraph{Dataset design.} 
We focus on the scenarios where objects have similar or even the same appearance and diverse motion patterns, including frequent crossover, occlusion and body deformation. The first property makes tracking by purely comparing object appearance invalid because the extracted visual features are no longer distinguishable for different objects. The second property further requires more informative clues rather than appearance in tracking, such as motion analysis and temporal dynamics. 

We argue that ``crowd’’ by simply increasing the density of objects is not what we expect. For example, MOT20~\cite{MOT20} contains videos where groups of pedestrians are very crowded. But as the pedestrian movement is very regular, the relative position and occlusion area keep almost consistent, such ``crowd’’ is not an obstacle for appearance matching. Therefore, we focus on situations where multiple objects are moving in a ``relatively'' large range, where the occluded areas are dynamically changing, and they are even in crossover. Such cases are common in real world but naive linear motion models can not handle them anymore.

\begin{table}[t]
\begin{center}
{\setlength{\tabcolsep}{1.5mm}
\begin{tabular}{l | l l l}
\toprule
Dataset & MOT17~\cite{MOT16} & MOT20~\cite{MOT20} & DanceTrack \\
\midrule
Videos & 14 & 8 & \textbf{100} \\
Avg. tracks & 96 & \textbf{432} & 9\\
Total tracks & 1342 & \textbf{3456} & 990\\
Avg. len. (s) & 35.4 & \textbf{66.8} & 52.9 \\
Total len. (s) & 463 & 535 & \textbf{5292}\\
FPS & \textbf{30} & 25 & 20 \\
Total images & 11,235 & 13,410 & \textbf{105,855}\\
\bottomrule
\end{tabular}}
\end{center}
\vspace{-3mm}
\caption{The comparison of dataset meta-information between DanceTrack and its closest benchmark for multi-human tracking, MOT17 and MOT20. DanceTrack contains much more videos and images than MOT datasets.}
\label{table_video}
\vspace{-5mm}
\end{table}

\begin{figure*}
    \centering
    \includegraphics[width=\linewidth]{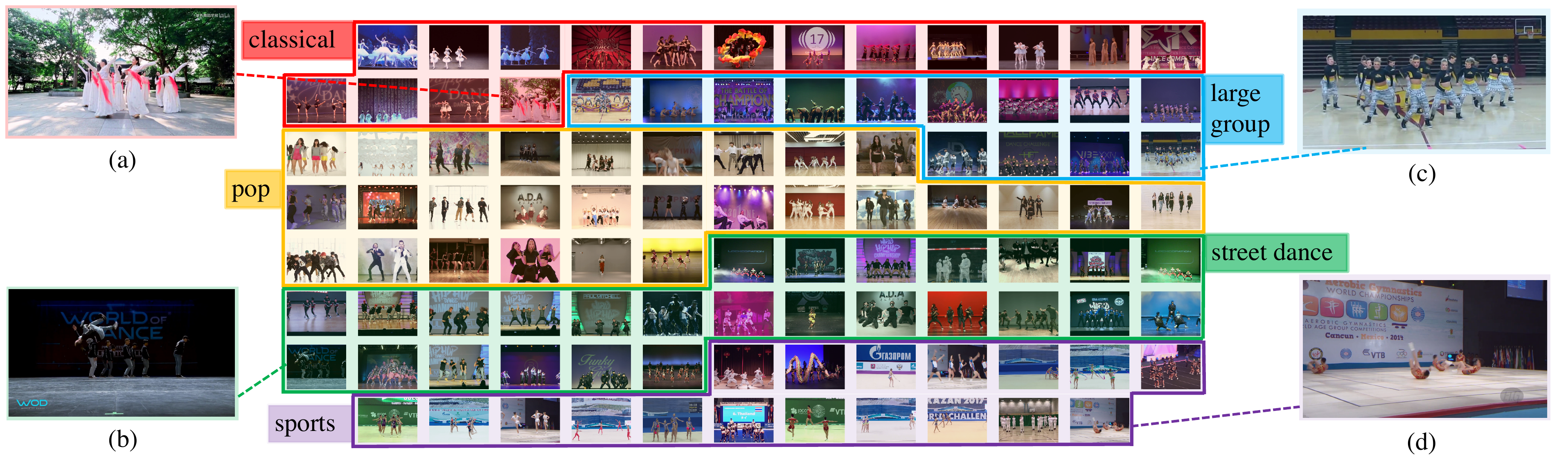}
    \caption{Sampled scenes from DanceTrack dataset.  DanceTrack contains multiple genres of dance, including classical dance, street dance, pop dance, large group dance and sports. The scenes in DanceTrack are diverse: (a) outdoor scenes; (b) low-lighting and distant camera scenes; (c) large group of dancing people; (d) gymnastics scene where the motion is usually even more diverse and people have more aggressive deformation.}
  \label{figure:scene}
  \vspace{-2mm}
\end{figure*}

\myparagraph{Video collection.} 
To achieve the design goals described above, we collected videos including mostly group dancing from the Internet. As shown in Figure~\ref{figure:scene}, the dancers usually wear very similar or even the same clothes. They make a large-range motion, diverse gestures and frequent crossover. These properties greatly satisfy our motivation. We collect the videos from different search engines with keywords like ``street dance’’, ``hip-pop dance’’, ``cheerleading dance’’, ``rhythmic gymnastics’’ and so on. The collection is only for publicly available videos and under the permit of fair use of video resources.

\myparagraph{Annotation.} 
We use a commercial tool to annotate the collected videos. The annotated labels include bounding boxes and identifications. For a partly-occluded object, a full-body box is annotated. For a fully-occluded object, we do not annotate it; when it re-appears in the future frame, its identification is kept as the same as in the previous frame when it is visible. To facilitate the annotation process, our tool can automatically propagate the annotated boxes from the previous frame to the current frame, and the annotator only needs to refine the boxes in the current frame. To build a high-quality dataset, the annotations have been checked by another group of people and errors are reported back to the annotators for re-annotation.

\subsection{Dataset Statistic}
\label{sec:dataset_stat}
We provide some analytical information of DanceTrack dataset and compare it with existing multi-object tracking datasets. The statistical information helps to understand the uniqueness of the proposed dataset.

\myparagraph{Dataset split.}
 We collect 100 videos in DanceTrack dataset, by default using 40 videos as training set, 25 as validation set and 35 as test set. For splitting, we keep the distribution of subsets close in terms of average length, average bounding box number, scenes and the motion diversity. We make the annotation of training set and validation set public while keeping the testing set annotation private for competition use. Some basic information of DanceTrack is shown in Table~\ref{table_video}.
Compared with MOT datasets, DanceTrack has much larger volume (10x more images and 10x more videos). MOT20 focuses on crowded scenes, so it has more tracks but the appearance of objects is very distinguishable and their motion is regular. As a consequence, the association on MOT20 still requires little motion estimation when good detection results are provided.

\vspace{-1mm}
\myparagraph{Scene diversity.}
DanceTrack contains diverse scenes. Samples from all 100 videos are provided in Figure~\ref{figure:scene}. One shared property for all videos is that the instances of people in a video usually have very similar appearance. This is designed on purpose to avoid the shortcut of tracking by pure appearance matching. DanceTrack contains multiple genres of dance, such as street dance, pop dance, classical dance (ballet, tango, etc.) and large group dancing. It also contains some sports scenarios such as gymnastics, Chinese Kung Fu and cheerleader dancing. Figure~\ref{figure:scene}(a) shows outdoor scenes though most included videos are indoor. Figure~\ref{figure:scene}(b) shows some especially hard cases, such as low lighting and distant camera. Figure~\ref{figure:scene}(c) shows a large group of people dancing, including at most 40 people. Figure~\ref{figure:scene}(d) shows gymnastics where people show extremely diverse body gestures, frequent pose variation and complicated motion pattern.

\vspace{-1mm}
\myparagraph{Appearance similarity.}
We make quantitative analysis about how appearance-only matching is not reliable on DanceTrack by measuring the appearance similarity among objects. We use a pre-trained re-ID model~\cite{deepsort_pytorch} to extract the appearance features $F(B_i^t)$ of object $B_i$ on a frame $t$, and then compute the sum of cosine distance of the re-ID features among objects in the video as
\begin{equation}
    V = \frac{1}{T} \sum_{t=1}^T \frac{1}{N_t^2}\sum_{i}^{N_t} \sum_{j \neq i}^{N_t}( 1 - \texttt{cos}<F(B_i^t), F(B_j^t)>),
\end{equation}
where $T$ is the number of frames in the video sequence, $N_t$ is the number of objects on the frame $t$ and $<$$\cdot$$>$ is the angle between two vectors.

\begin{figure*}
\begin{minipage}{1.0\columnwidth}
    \subfloat[Cosine distance of re-ID feature]{ 
    \label{a} 
    \includegraphics[width=0.95\linewidth]{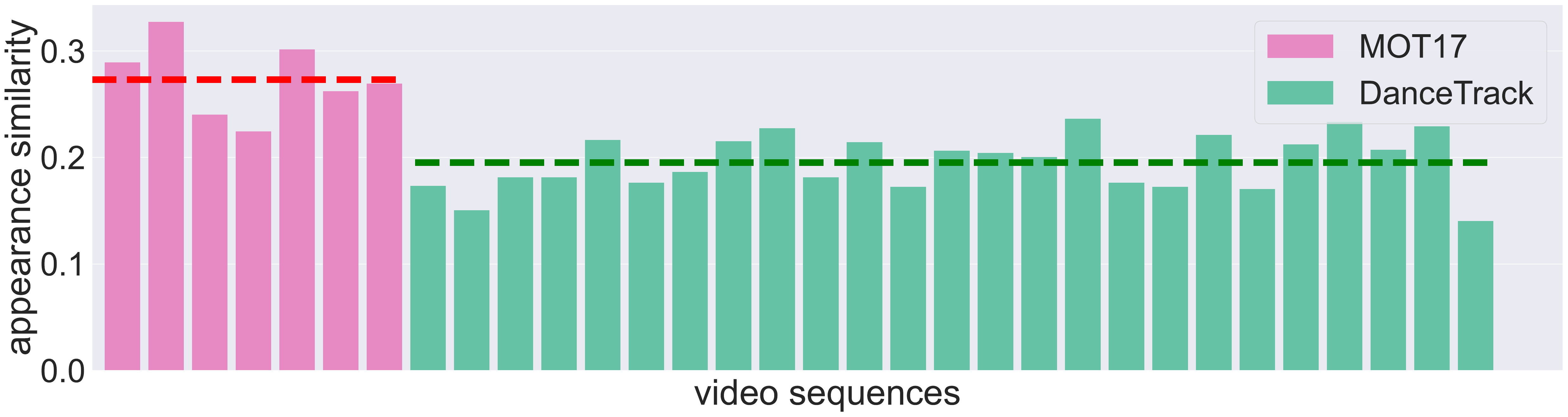}
    }
    \subfloat[IoU on adjacent frames]{ 
    \label{b} 
    \includegraphics[width=.525\linewidth]{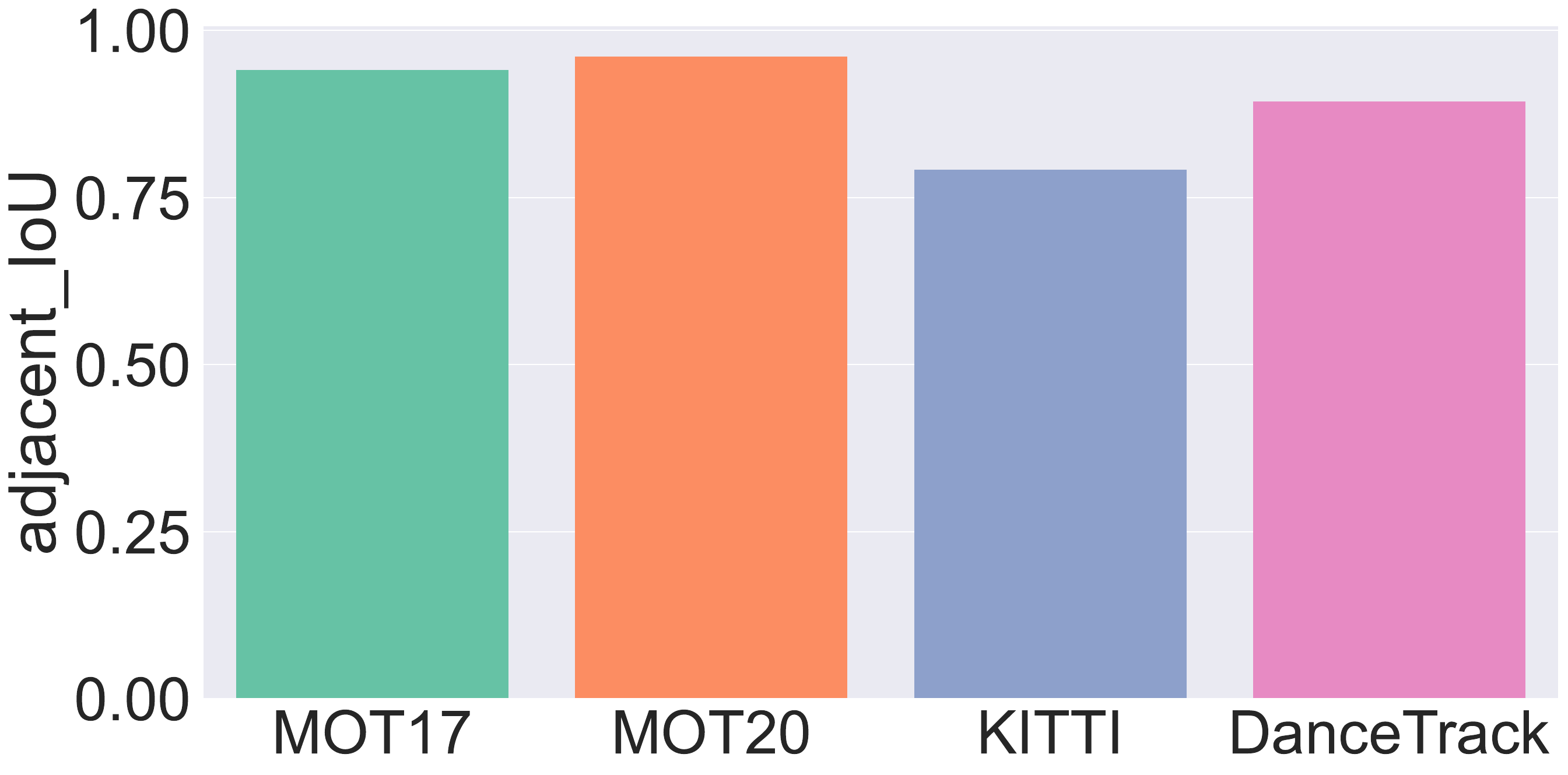}
    }
    \subfloat[Frequency of relative position switch]{ 
    \label{c} 
    \includegraphics[width=.525\linewidth]{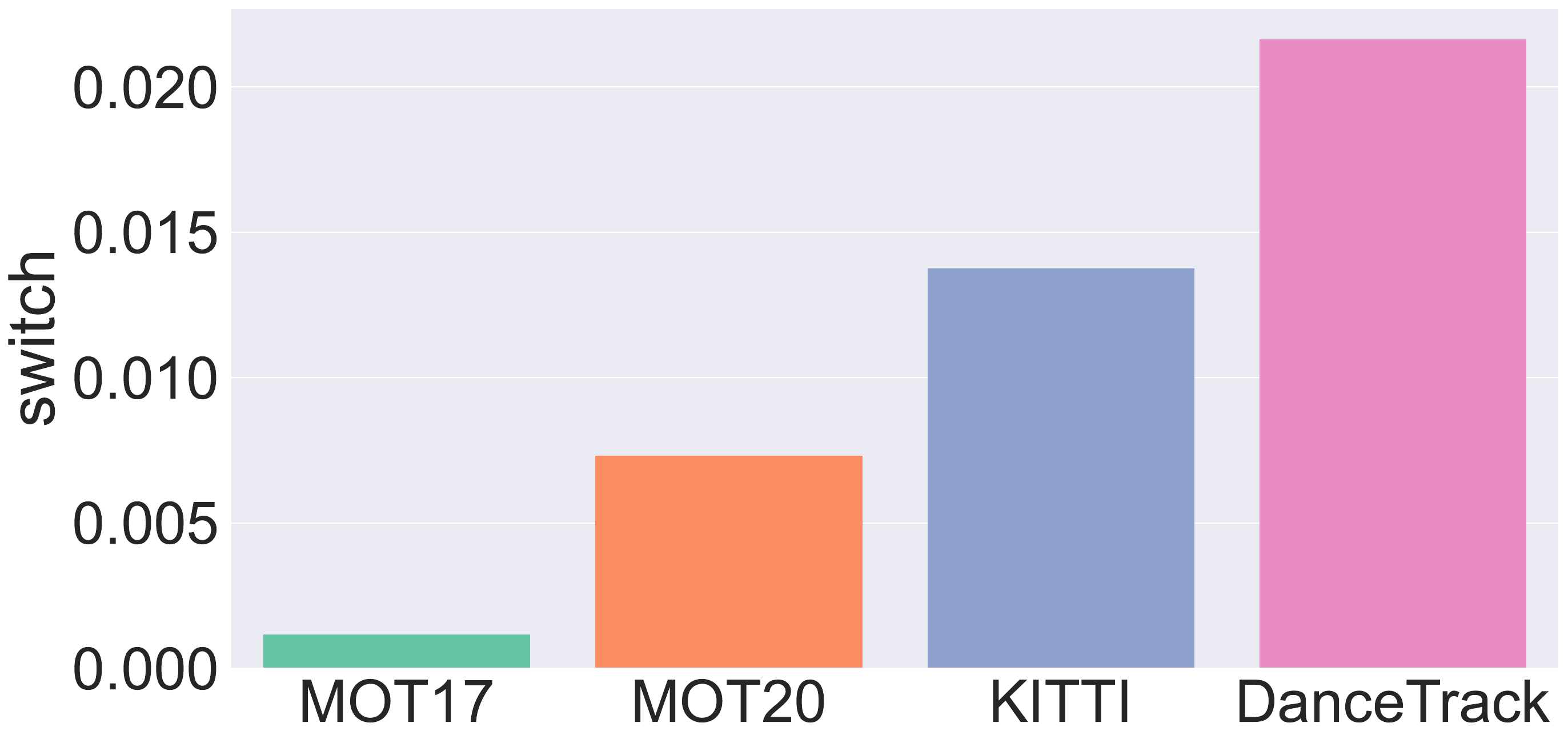}
    }
\end{minipage}

    \caption{(a) Cosine distance of re-ID features. The dashed lines are for the average cosine distance similarity for the two datasets. The cosine distance of re-ID features of DanceTrack is  lower than that of MOT17, in other words, the appearance similarity between different objects is higher. (b) IoU on adjacent frames. Compared to MOT17 and MOT20, DanceTrack has a similar score. It means that the frame rate and object motion speed are still reasonable in DanceTrack. 
    (c) Frequency of relative position switch. This metric measures the frequency of crossover and is highly related to the occlusion between objects. DanceTrack has much more frequent relative position switches than other pedestrian tracking datasets, such as MOT17 and MOT20. Even compared to the driving dataset KITTI, where the moving camera naturally causes many relative position switches, DanceTrack still has a higher frequency.}
    \label{fig:quant}
\end{figure*}

We compare the object appearance similarity in DanceTrack to that in MOT17 dataset, as shown in Figure~\ref{fig:quant}(a), each bin represents one video sequence. It is obvious that the cosine distance of re-ID features of DanceTrack is lower than that of MOT17, in other words, the appearance similarity among co-existing objects is higher. This quantitative analysis shows the challenge of DanceTrack to current popular appearance matching for association.

\myparagraph{Motion pattern.} 
We introduce two metrics to analyze the motion pattern in DanceTrack dataset and compare that to other multi-object tracking datasets.

\noindent \textit{IoU on adjacent frames}: a natural measurement of object movement range is its bounding-box IoU (Intersection-over-Union) on two adjacent frames. A low IoU indicates fast-moving objects or the low frame rate of videos.
Given a video with $N$ objects and $T$ frames,
the averaged IoU on adjacent frames for this video is 
    \begin{equation}
        U = \frac{1}{N(T-1)} \sum_{i}^N \sum_{t=1}^{T-1} IoU(B_i^t, B_i^{t+1}).
    \end{equation}

\noindent \textit{Frequency of Relative Position Switch}: a metric to measure the diversity of objects' motion in a global view is the frequency for two objects to switch their relative position. This could happen between leftward and rightward or between upward and downward. On the contrary, movement with consistent velocity tends to cause a lower chance of relative position switch. Given a video, the average frequency of relative position switch is defined as 
    \begin{equation}
        S = \frac{\sum_{i}^N \sum_{j \neq i}^N \sum_{t=1}^{T-1} sw(B_i^t, B_j^t, B_i^{t+1}, B_j^{t+1})}{2N(T-1)(N-1)}, 
    \end{equation}
where $sw$ is an indicator function, where $sw($$\cdot$$)$=1 if the two objects swap their left-right relative position or top-down relative position on the adjacent frames, $sw($$\cdot$$)$=0 if there is no swap. We measure their relative position by comparing their bounding box center locations. And considering that such crossover causes potential difficulty only when the objects have overlap, we only take the objects with overlap into the calculation.

From the results shown in Figure~\ref{fig:quant}(b), we could find that DanceTrack and MOT datasets have close average IoU on adjacent frames. This indicates that DanceTrack 
does not have unreasonably fast object movement.

On the other hand, from Figure~\ref{fig:quant}(c) we could find that DanceTrack has much more frequent relative position switches than other datasets such as KITTI, MOT17 and MOT20. 
The frequent relative position switches are caused by highly non-linear motion pattern and result in frequent crossover and inter-object occlusion. This result shows that the challenge of motion diversity in DanceTrack.

\begin{table*}[t]
\begin{center}
{\setlength{\tabcolsep}{2.0mm}

\begin{tabular}{c c c |c c c c c | c c c c c}

\arrayrulecolor{white}\hline
\Xhline{2\arrayrulewidth}
\arrayrulecolor{white}\hline

\arrayrulecolor{black}
\multirow{2}{*}{Appearance} & 
\multirow{2}{*}{IoU} & 
\multirow{2}{*}{Motion} & 
\multicolumn{5}{c|}{MOT17}  & 
\multicolumn{5}{c}{DanceTrack (Proposed Dataset)} \\
\arrayrulecolor{black}\cline{4-8} \cline{9-13}
& & & HOTA & DetA & AssA & MOTA & IDF1 &  
  HOTA & DetA & AssA & MOTA & IDF1  \\
\arrayrulecolor{white}\hline
\arrayrulecolor{black}\hline
\arrayrulecolor{white}\hline
& \checkmark &  & \textbf{98.1} & 98.9 & \textbf{97.3} & 98.0& 97.8 & \textbf{72.8} & \textbf{98.9} & 53.6 & 98.7 & 63.5\\
& \checkmark & \checkmark & 96.4 & 97.1 & 95.8 & \textbf{99.7} & 98.1 & 69.4 & 87.9 & \textbf{54.8} & \textbf{99.4} & \textbf{71.3}\\
\checkmark & \checkmark & \checkmark & 95.0 & 94.7 & 95.4 & 99.3 & \textbf{98.8} & 59.7 & 82.5 & 43.2 & 97.2 & 60.5\\
\checkmark & &  & 93.3 & \textbf{99.0} & 87.9 & 98.9 & 90.9 & 68.0 & 97.7 & 47.4 & 97.9 & 58.7\\
\arrayrulecolor{white}\hline
\Xhline{2\arrayrulewidth}
\arrayrulecolor{white}\hline
\end{tabular}

}
\end{center}
\vspace{-2mm}
\caption{Oracle analysis of different association models on MOT17 and DanceTrack validation set, respectively. The detection boxes are ground-truth boxes. The result comparison shows the evident increased difficulty of performing multi-object tracking on DanceTrack than MOT17 dataset.
}
\label{table_oracle}
\vspace{-8mm}
\end{table*}

\subsection{Evaluation Metrics}
\label{sec:metrics}
\vspace{-0.2cm}
For a long time, multi-object tracking community used Multi-Object Tracking Accuracy (MOTA) as the main metric for evaluation. However, recently, the community realizes that MOTA focuses too much on detection quality instead of association quality. Thus, Higher Order Tracking Accuracy (HOTA)~\cite{HOTA} is proposed to correct this historical bias. Up to now, HOTA has been used for the main metrics to evaluate tracking quality on multiple popular benchmarks such as BDD100K~\cite{BDD} and KITTI~\cite{KITTI}. We follow this setting for evaluation metrics of DanceTrack. In our protocol, the main metric is HOTA. We also use AssA and IDF1 score to measure association performance and DetA and MOTA for detection quality.
For the detailed definitions of these metrics, we refer to ~\cite{mota,idf1,HOTA}. To make it convenient to run for fine-grained analysis, the evaluation tools also provide previously widely-used statistics, such as False Positive (FP), False Negative (FN) and ID switch (IDs).

\subsection{Limitation}
We discuss some limitations of the proposed dataset. 
 First, given the mentioned motivation and the proposed dataset, we do not provide an algorithm that highly outperforms previous multi-object tracking algorithms but keep this as an open question for future study. Second, for the cases we emphasize in this work, the annotation of human pose or segmentation mask should be important for more fine-grained study. But limited by time and resources, we only provide the annotation of bounding boxes in this version.

\vspace{-0.2cm}
\section{Experiments}
\subsection{Experiment Setup}
\vspace{-0.2cm}
\myparagraph{Dataset configurations} We compare DanceTrack with its closest dataset, MOT17. For MOT17, because the test server is not available easily, we follow the train-val splitting provided in CenterTrack~\cite{CenterNet} to evaluate on the validation subset, unless in Section~\ref{sec:sota}. For DanceTrack, we follow the default splitting described in the previous section.

\myparagraph{Model configuration} Unless specified otherwise, we inherit the default training settings of the investigated algorithms provided in the original papers or the officially released codebases. 

\subsection{Oracle Analysis}
\label{sec:oracle}
\vspace{-0.1cm}
To decompose the analysis over object localization and association, we perform oracle analysis here. We use the ground truth bounding boxes with different association algorithms to achieve the upper-bound performance. This analysis can help us to understand what is the true bottleneck of tracking on different datasets. 

We compare IoU matching, motion modeling and appearance matching for the association. IoU matching is simply performed by calculating the IoU of objects' bounding boxes in adjacent frames.  We use a pre-trained Re-ID model~\cite{deepsort_pytorch} for appearance matching and a Kalman Filter~\cite{kalman1960new} for motion modeling under linear motion assumption. We have experiments on MOT17 and DanceTrack respectively. The results are shown in Table~\ref{table_oracle}.

From the results, the performance is almost perfect in terms of all metrics on MOT17. 
And it is interesting that using only IoU matching achieves the best performance, which proves that MOT17 contains objects with simple and regular motion patterns and the bottleneck does not lie in association in most cases.  

\begin{figure}[t]
    \centering
    \includegraphics[width=0.95\linewidth]{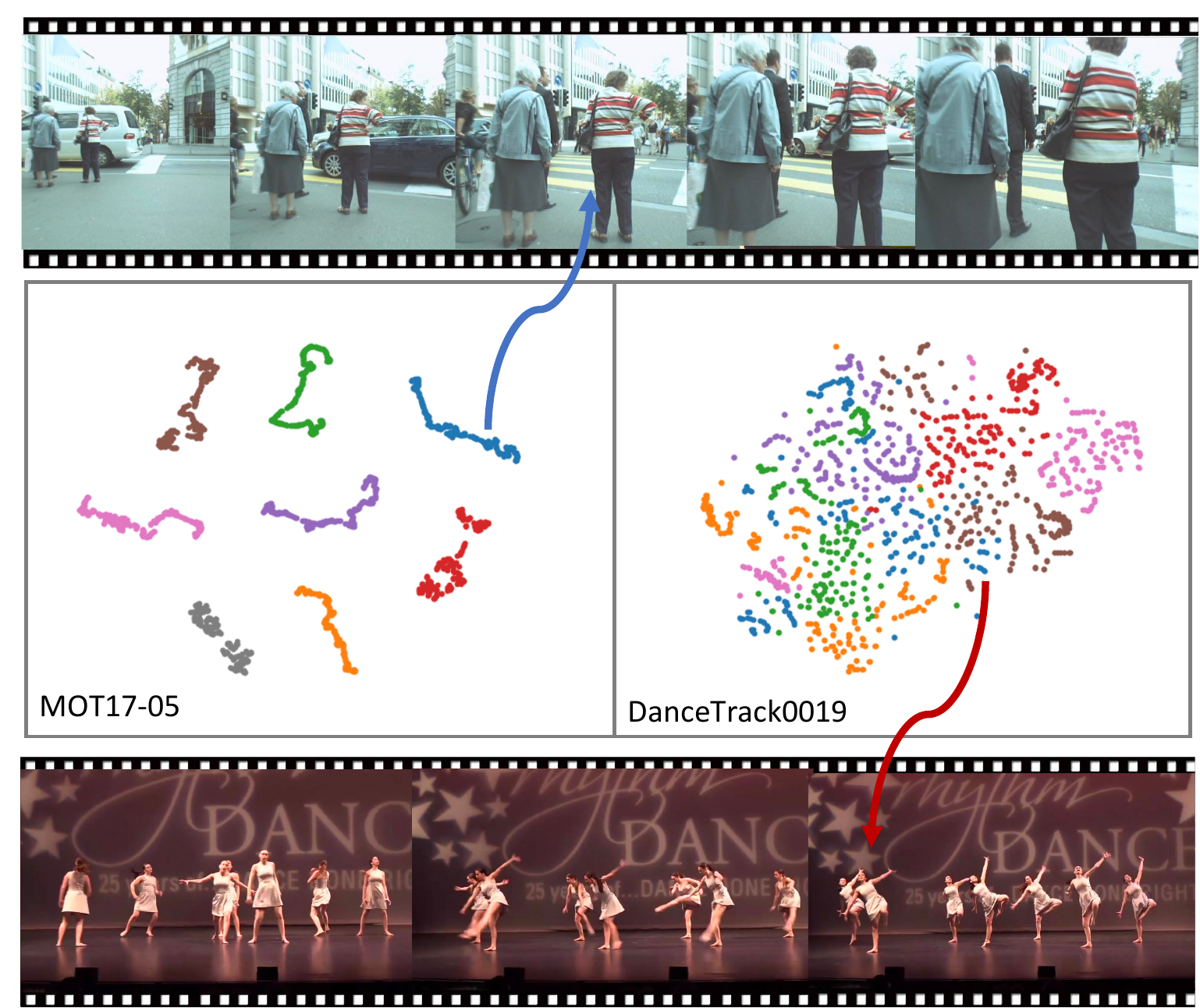}
    \caption{Visualization of re-ID feature from sampled video in MOT17 and DanceTrack dataset using t-SNE~\cite{tsne}. The same object is coded by the same color. For better visualization, we only select first 200 frames in each video sequence.}
  \label{figure:tsne}
\vspace{-2mm}
\end{figure}

On the other hand, using only IoU matching on DanceTrack gives a much lower performance than on MOT17. Given DetA and MOTA scores are already close to 100, the bottleneck is obviously in the association part. All association metric scores in all cases experience a dramatic drop compared with that on MOT17. Besides, the best performance lies in only IoU matching, even combining a linear motion model or additional appearance information does not help. When using appearance similarity, all metrics are worse than not using any appearance cue. This is because the objects in DanceTrack videos usually have indistinguishable appearance so simply using appearance matching makes negative effects in some cases. In Figure~\ref{figure:tsne}, we visualize the appearance feature of objects extracted from DanceTrack and MOT17 videos respectively. 
We can observe that the appearance features of different objects are very distinguishable in the feature space on MOT17 while highly entangled on DanceTrack. This qualitatively provides evidence for the high similar appearance of objects in the proposed DanceTrack dataset. 

Given the results shown in the analysis with oracle object localization, we can reach a clear conclusion that existing datasets have a heavy bias that it focuses more on the detection quality only and the involved simple trajectory patterns limit the study in this area. On the contrary, DanceTrack is proposing a much higher requirement to develop multi-object trackers with improvement in association ability. Considering the scenarios included in DanceTrack are what we experience in real life, we believe it is meaningful to provide such a platform.

\begin{table*}[!t]
\begin{center}
{\setlength{\tabcolsep}{3.5mm}

\begin{tabular}{l |p{7mm} p{7mm}p{7mm}p{7mm}p{7mm} | p{7mm}p{7mm}p{7mm}p{7mm}p{7mm}}

\arrayrulecolor{white}\hline
\Xhline{2\arrayrulewidth}
\arrayrulecolor{white}\hline

\arrayrulecolor{black}
\multirow{2}{*}{Methods} & \multicolumn{5}{c|}{MOT17}  & \multicolumn{5}{c}{DanceTrack (Proposed Dataset)} \\
\arrayrulecolor{black}\cline{2-6} \cline{7-11}
& HOTA & DetA & AssA & MOTA & IDF1 &
  HOTA & DetA & AssA & MOTA & IDF1\\
\arrayrulecolor{black}\hline
CenterTrack$^1$~\cite{CenterTrack} & 52.2 & 53.8 & 51.0 & 67.8 & 64.7 & 41.8 & 78.1 & 22.6 & 86.8 & 35.7\\
FairMOT$^1$~\cite{FairMOT} & 59.3 & 60.9 & 58.0 & 73.7 & 72.3 & 39.7 & 66.7 & 23.8 & 82.2 & 40.8 \\
QDTrack$^3$~\cite{quasidense} & 53.9 & 55.6 & 52.7 & 68.7 & 66.3& 54.2 & 80.1 & 36.8 & 87.7 & 50.4\\
TransTrack$^1$~\cite{Transtrack} &  54.1 & 61.6 & 47.9 & 75.2 & 63.5 & 45.5 & 75.9 & 27.5 & 88.4 & 45.2\\
TraDes$^1$~\cite{TraDeS} & 52.7 & 55.2 & 50.8 & 69.1 & 63.9 & 43.3 & 74.5 & 25.4 & 86.2 & 41.2\\
MOTR$^2$~\cite{MOTR} & 57.2 & 58.9 & 55.8 & 71.9 & 68.4 & 54.2 & 73.5 & \textbf{40.2}	& 79.7 & 51.5 \\
GTR$^2$~\cite{zhou2022global} & 59.1 & 61.6 & 57.0 & 75.3 & 71.5 & 48.0 & 72.5 & 31.9 & 84.7 &  50.3\\
ByteTrack$^1$~\cite{bytetrack}& 63.1 & \textbf{64.5} & 62.0 & \textbf{80.3} & 77.3 &  47.7 & 71.0 & 32.1 & 89.6 & 53.9\\
OC-SORT$^2$~\cite{cao2022observation} & \textbf{63.2} & 63.2 & \textbf{63.2} & 78.0 & \textbf{77.5} & \textbf{55.1} & \textbf{80.3} & 38.3 & \textbf{92.0} & \textbf{54.6}\\
\arrayrulecolor{white}\hline
\Xhline{2\arrayrulewidth}
\arrayrulecolor{white}\hline

\end{tabular}
}
\end{center}
\vspace{-3mm}
\caption{Tracking performance of investigated algorithms on MOT17 and DanceTrack \textbf{test set}. The result comparison shows the evident increased difficulty of performing multi-object tracking on DanceTrack than MOT17 dataset.
}
\label{table_mot}
\vspace{-8mm}
\end{table*}

\subsection{Benchmark Results}
\label{sec:sota}
\vspace{-0.1cm}
We benchmark the current state-of-the-art multi-object tracking algorithms on MOT17 and DanceTrack. The evaluation is in the ``private''  setting that the algorithm performs both detection and association. The methods basically inherit the settings on MOT17 for training on DanceTrack training set. The benchmark results are reported in Table~\ref{table_mot}. 

For tracking quality measured by HOTA, IDF1 and AssA, all algorithms show a significant performance gap from MOT17 to DanceTrack. For all investigated methods, their performance on DanceTrack is far from satisfactory. Notably, the detection quality metrics, MOTA and DetA, of all algorithms are in fact higher on DanceTrack than on MOT17. This suggests that detection is not the bottleneck to have good tracking performance on DanceTrack and further highlights the drop of association performance. The benchmark results prove that DanceTrack raises the challenge to make robust association in the cases of the uniform appearance and the diverse motion of objects.

\begin{table}[t]
\begin{center}
{\setlength{\tabcolsep}{1.3mm}
\begin{tabular}{l c c c c c}
\toprule
Association & HOTA & DetA & AssA & MOTA & IDF1\\
\midrule
IoU & 44.7 & 79.6 & 25.3& 87.3& 36.8\\
SORT\cite{SORT} & 47.8 & 74.0 & 31.0 & \textbf{88.2} & 48.3\\ 
DeepSORT\cite{DeepSORT} & 45.8 & 70.9 & 29.7 & 87.1 & 46.8\\
MOTDT\cite{MOTDT} & 39.2 & 68.8 & 22.5 & 84.3 & 39.6\\
BYTE\cite{bytetrack}& 47.1 & 70.5 & 31.5 & \textbf{88.2} & \textbf{51.9}\\
OC-SORT\cite{cao2022observation} & \textbf{52.1} & \textbf{79.8} & \textbf{35.3} & 87.3 & 51.6 \\
\bottomrule
\end{tabular}
}
\end{center}
\vspace{-3mm}
\caption{Comparison of different association algorithms on DanceTrack validation set. The detection results are output by YOLOX~\cite{YOLOX} detector, trained on DanceTrack training set.}
\label{table_tracker}
\vspace{-6mm}
\end{table}

\begin{table}[t]
\begin{center}
{\setlength{\tabcolsep}{1.2mm}
\begin{tabular}{l c c c c c}
\toprule
Motion & HOTA & DetA & AssA & MOTA & IDF1\\
\midrule
None(IoU) & 44.7 & \textbf{79.6} & 25.3 & 87.3 & 36.8\\
Kalman filter\cite{SORT} & 47.8 & 74.0 & 31.0 & 88.2 & 48.3\\ 
LSTM\cite{DEFT} & \textbf{51.6} & 78.2 & \textbf{34.2} & \textbf{89.2} & \textbf{50.8}  \\
\bottomrule
\end{tabular}}
\end{center}
\vspace{-3mm}
\caption{Comparison of different motion models on DanceTrack validation set. The detection results are output by YOLOX~\cite{YOLOX} detector, trained on DanceTrack training set.}
\label{table_motion}
\vspace{-8mm}
\end{table}

\subsection{Association Strategy}
\vspace{-0.1cm}
The methods in the previous section entangle the detection and tracking modules. To have an independent study on association algorithms, we use the most recently developed YOLOX~\cite{YOLOX} detector for object detection on DanceTrack and conduct different association algorithms following that. The results are shown in Table~\ref{table_tracker}.

\footnotetext{The number is arXiv version. For details, refer to change log in appendix. We thank the authors for reporting their updated results.}
SORT~\cite{SORT} uses Kalman Filter to model the object motion and DeepSORT~\cite{DeepSORT} adds appearance matching. Compared to SORT, DeepSORT shows no performance boost but worse performance instead, suggesting the negative gain due to appearance matching. On the other hand, MOTDT~\cite{MOTDT} uses the tracking result to help detect bounding boxes. But in fact, detection performance can be really good on DanceTrack dataset and the exact bottleneck is the association part, so MOTDT shows even worse performance on both detection quality and association quality with its design. Lastly, BYTE~\cite{bytetrack} uses a high-tolerance strategy to select detection results into the association stage. The design aims to decrease tracklet fragmentation in tracking. OC-SORT~\cite{cao2022observation,contributors2020mmtracking} improves the association robustness for non-linear object motion in pure motion-based manner and it shows the best association performance on DanceTrack where object motion is highly non-linear. The results reveal that DanceTrack is not a strict challenge for object detectors, the true challenge is in the object association part. 

We further use different motion models to introduce temporal dynamics in the tracking process to facilitate better association, as shown in Table~\ref{table_motion}. Obviously, both Kalman filter~\cite{SORT} and LSTM~\cite{DEFT} outperform naive IoU association (without temporal dynamics) by a large margin, indicating the great potential of motion models in tracking objects, especially when appearance cues are not reliable. With the relatively slow progress of object model motion in the field of multi-object tracking, we expect to see more researches.

\subsection{Analysis of More Modalities}
\vspace{-0.1cm}
\label{modalities}
Considering high scores of MOTA and DetA on DanceTrack, the limited performance on DanceTrack is an exact failure of trackers instead of detectors. To boost performance, a straightforward strategy is to add more cues other than frame-wise bounding box. Since DanceTrack contains bounding boxes and identities annotations only, we propose to use joint-training technology with other datasets to enable the model output more modalities.

\begin{figure*}[!t]
\begin{center}
\includegraphics[width=0.95\linewidth]{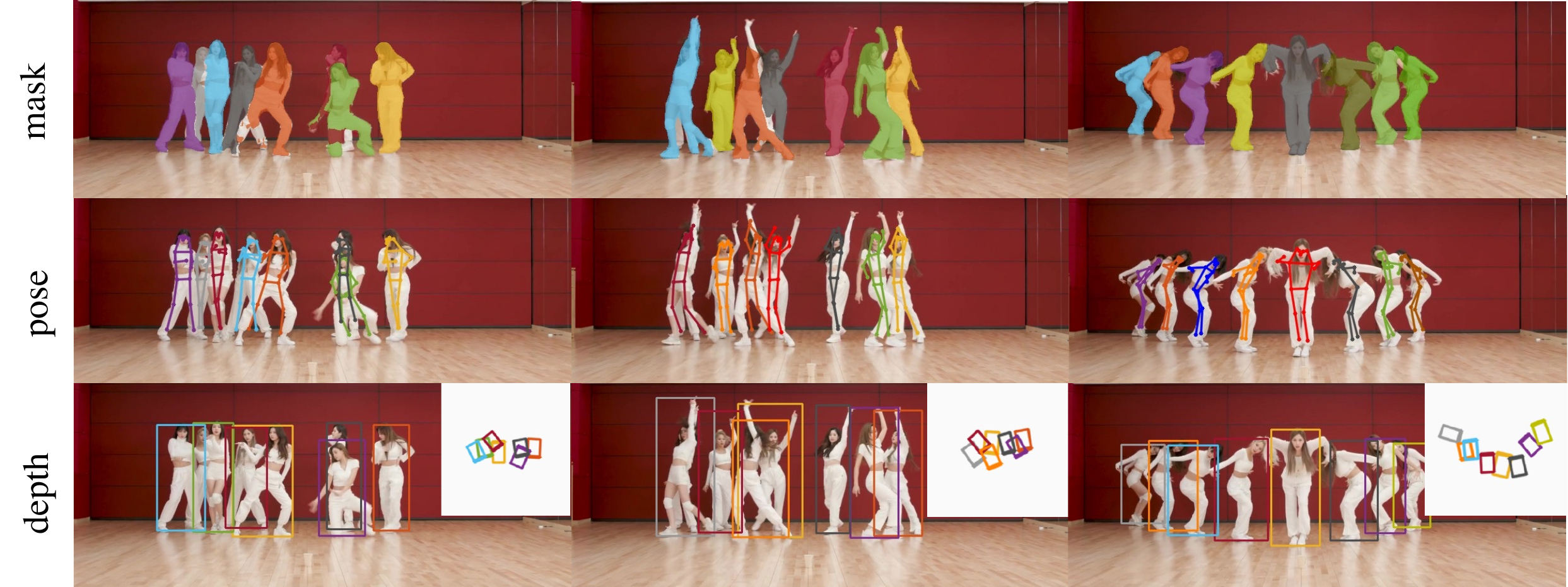}    
\end{center}
\vspace{-5mm}
\caption{Visualization of adding more information beyond bounding box on DanceTrack. Tracks are coded by color. The 1st, 2nd and 3rd column are frame20, 120 and 200 of DanceTrack0007 video.}
\label{figure:modal}
\end{figure*}

\begin{table*}[t]
\begin{center}
{\setlength{\tabcolsep}{4.0mm}
\begin{tabular}{l c |l l l l l}
\toprule
Data & Ass. & HOTA & DetA & AssA & MOTA & IDF1  \\
\midrule
DanceTrack & box & 36.9 & 63.6 & 21.6 & 78.8 & 39.2\\
+ COCOmask~\cite{COCO} & box & 38.1	\upscore{1.2} & 64.5 \upscore{0.9} & 22.6  \upscore{1.0} & 80.6 \upscore{1.8} & 40.3 \upscore{1.1} \\ 
+ COCOmask & + mask & 39.2 \upscore{1.1} & 
64.9 \upscore{0.4}	& 
23.9 \textcolor{codegreen}{(+\textbf{1.3})}& 
80.7 \upscore{0.1} & 
41.6 \textcolor{codegreen}{(+\textbf{0.3})}\\

\midrule
DanceTrack & box & 36.9 & 63.6 & 21.6 & 78.8 & 39.2\\
+ COCOpose~\cite{COCO} & box & 40.6 \upscore{3.7} & 65.5 \upscore{1.9} & 25.3 \upscore{3.7} & 82.9 \upscore{4.1} & 42.9 \upscore{3.7}\\ 
+ COCOpose & + pose & 
41.0 \upscore{0.4} & 
65.9 \upscore{0.4} & 
25.6 \upscore{0.3} & 
83.1 \upscore{0.3} &
43.9 \textcolor{codegreen}{(+\textbf{1.0})} \\

\midrule
DanceTrack & box & 36.9 & 63.6 & 21.6 & 78.8 & 39.2\\
+ KITTI~\cite{KITTI} & box & 34.4 \downscore{2.5} & 57.8	\downscore{5.8} & 20.7 \downscore{0.9} & 72.9 \downscore{5.9} & 38.5 \downscore{0.7} \\ 
+ KITTI & + depth  & 
35.1 \textcolor{codegreen}{(+\textbf{0.7})} & 
57.3 \downscore{0.5}& 
21.6 \textcolor{codegreen}{(+\textbf{0.9})} & 
72.8 \downscore{0.1}& 
40.2 \textcolor{codegreen}{(+\textbf{1.7})} \\
\bottomrule
\vspace{-0.3cm}
\end{tabular}}
\end{center}
\vspace{-3mm}
\caption{Ablation study on adding more information beyond bounding box on DanceTrack validation set. All experiments are based on CenterNet~\cite{CenterNet} model and BYTE~\cite{bytetrack} association.
(a) Segmentation mask improves the tracking performance on DanceTrack. (b) Pose information boosts the tracking performance with an even larger gap than segmentation mask. (c) Though adding depth information into association shows a slightly positive influence, the results still blame the domain shift between KITTI and DanceTrack.}
\label{table_mix}
\vspace{-6mm}
\end{table*}

\myparagraph{Does fine-grained representation help ?}
We investigate the influence of adding segmentation mask into the model. From Table~\ref{table_mix}, we observe a performance boost by using the segmentation mask. First, the introduction of more fine-grained annotation benefits the model by multi-task learning. Second, for crowded and occluded situations, mask is a more reliable information than bounding box to associate objects. Besides mask, adding pose information in training better boosts the model performance on DanceTrack, and using the output pose in association further helps to achieve better tracking results. When most areas of a human body are occluded, bounding box usually can not provide reliable output while the pose estimation model focusing on certain human body key-points usually shows higher robustness.

\myparagraph{Does depth information help ?} 
We use additional depth information to help tracking on DanceTrack. The results are shown in Table~\ref{table_mix}. In contrast to the COCO segmentation mask and human pose, depth information learned from KITTI dataset does not increase the performance on DanceTrack. We explain that COCO segmentation and pose estimation datasets contain human as the main category, while KITTI mainly contains vehicle instances. Thus, the object and scene prior in DanceTrack and KITTI change and this domain shift degenerates the model. Nevertheless, depth information indeed helps association performance if we regard the baseline as the model trained on joint-dataset of DanceTrack and KITTI.
However, limited by the available resources of depth-annotated data, this is the best we could try for now. We expect more study on the influence of depth information to associate objects with uniform appearance and diverse motion.

\vspace{-0.1cm}
\section{Conclusion}
\vspace{-0.2cm}
In this paper, we propose a new multi-object tracking dataset called DanceTrack. The objects have uniform appearance and diverse motion pattern in DanceTrack, preventing being taken short-cuts by Re-ID algorithms. The motivation behind it is to reveal the bias in existing datasets that tend to emphasize detection quality and matching appearance only. This makes other cues to associate objects underrepresented. We believe that the ability to analyze the complex motion pattern is necessary for building a more comprehensive and intelligent tracker. DanceTrack provides such a platform to encourage future works.

\myparagraph{Acknowledgement}
We would like to thank the annotator teams and coordinators to build DanceTrack dataset. We appreciate Xinshuo Weng, Yifu Zhang for valuable discussion and suggestions. We would also like to thank Vivek Roy, Pedro Morgado, Shuyang Sun for their proof reading and suggestions on paper writing. This work was sponsored in part by NSF NRI Award IIS2024173. Ping Luo is supported by the General Research Fund of HK No.27208720 and 17212120.
\newpage
{\small
\bibliographystyle{ieee_fullname}
\bibliography{egbib}
}

\appendix
\section{Change Log}
\myparagraph{Version 1 (2021-11-29)}

- Initial arXiv version.

\myparagraph{Version 2 (2022-05-05)}

- Update MOTR in Table 3. For more details, please refer to
\url{https://github.com/megvii-model/MOTR}. 

- Add GTR and OC-SORT in Table 3.

- Compare different motion models by using YOLOX detector, in new Table 5, to be consistent with Table 4.

\myparagraph{Version 3 (2022-05-24)}

- Update QDTrack in Table 3.
New QDTrack replaces the Faster RCNN detector in the original QDTrack with a YOLOX model, by applying the same data augmentation strategy as in ByteTrack, and trained on DanceTrack train for 24 epochs. The other hyperparameters are aligned with ByteTrack, e.g. confidence thresholds for new tracks and existing tracks, input image resolution. Neither the tracking algorithm was changed nor the embedding learning component. 

\end{document}